\documentclass{bmvc2k}
\pdfoutput=1
\usepackage{fixltx2e}
\usepackage{multirow}
\usepackage{makecell}

\usepackage{caption}
\definecolor{captioncolor}{rgb}{0,0,.4}
\DeclareCaptionFont{captioncolor}{\color{captioncolor}}
\usepackage{subfigure}
\usepackage{pifont}
\usepackage[table,x11names]{xcolor}
\usepackage{array}
\usepackage{amssymb}

\newcolumntype{L}[1]{>{\raggedright\let\newline\\\arraybackslash\hspace{0pt}\vspace{0pt}}m{#1}}
\newcolumntype{C}[1]{>{\centering\let\newline\\\arraybackslash\hspace{0pt}\vspace{0pt}}m{#1}}
\newcolumntype{R}[1]{>{\raggedleft\let\newline\\\arraybackslash\hspace{0pt}\vspace{0pt}}m{#1}}
\newcolumntype{V}[1]{>{\centering\let\newline\\\arraybackslash\hspace{5pt}\vspace{5pt}}m{#1}}

\newcommand{\cmark}{\ding{51}}%
\newcommand{\xmark}{\ding{55}}%

\usepackage{amsthm}
\theoremstyle{definition}
\newtheorem{definition}{Definition}[section]

\title{Dominant Set Clustering and Pooling for Multi-View 3D Object Recognition.}

\addauthor{Chu Wang}{http://www.cim.mcgill.ca/~chuwang/}{1}
\addauthor{Marcello Pelillo}{http://www.dsi.unive.it/~pelillo/}{2}
\addauthor{Kaleem Siddiqi}{http://www.cim.mcgill.ca/~siddiqi/}{1}

\addinstitution{
 School of Computer Science\\
 McGill University\\
 Montr\'eal, Canada
}
\addinstitution{
 DAIS / ECLT\\
 Ca' Foscari University of Venice\\
 Italy
}
\runninghead{Wang, Pelillo, Siddiqi}{Recurrent Cluster-pooling CNN}

\def\etal{\emph{et al}\bmvaOneDot}

\begin{document}

\maketitle
\begin{abstract}
View based strategies for 3D object recognition have proven to be very successful. The state-of-the-art methods now achieve over 90\% correct category level recognition performance on appearance images. We improve upon these methods by introducing a view clustering and pooling layer based on {\em dominant sets}. The key idea is to pool information from views which are similar and thus belong to the same cluster. The pooled feature vectors are then fed as inputs to the same layer, in a recurrent fashion. This recurrent clustering and pooling module, when inserted in an off-the-shelf pretrained CNN, boosts performance for multi-view 3D object recognition, achieving a new state of the art test set recognition accuracy of 93.8\% on the ModelNet 40 database. We also explore a fast approximate learning strategy for our cluster-pooling CNN, which, while sacrificing end-to-end learning, greatly improves its training efficiency with only a slight reduction of recognition accuracy to 93.3\%. Our implementation is available at \url{https://github.com/fate3439/dscnn}.

\end{abstract}

\section{Introduction}
Appearance based object recognition remains a fundamental challenge to computer vision systems. Recent strategies have focused largely on learning category level object labels from 2D features obtained from projection. With the parallel advance in 3D sensing technologies, such as the Kinect, we also have the real possibility to seamless include features derived from 3D shape into recognition pipelines. There is also a growing interest in 3D shape recognition from databases of 3D mesh models, acquired from computer graphics databases \cite{modelnet} or reconstructed from point cloud data \cite{qi2016volumetric}. 

3D object recognition from shape features may be categorized into \textit{view-based} versus \textit{volumetric} approaches. View based approaches including those in \cite{chen2003visual, lowe2001local, macrini2002view, murase1995visual} create hand-designed feature descriptors from 2D renderings of a 3D object by combing information across different views. 3D object recognition is then reduced to a classification problem on the designed feature descriptors. More recent methods in this class combine CNN features from the multiple views to boost object recognition accuracy \cite{johns2016pairwise, qi2016volumetric, su2015multi,bai2016gift}. Volumetric approaches rely on 3D features computed directly from native 3D representations, including meshes, voxelized 3D grids and point clouds \cite{horn1984extended,kazhdan2003rotation,knopp2010hough,sinha2016deep}. The state-of-the-art here is the use of 3D convolutional neural networks on discretized occupancy grids for feature extraction for further processing or for direct classification \cite{maturana2015voxnet, qi2016volumetric, wu20153d,wu2016learning}. At the present time, at least when evaluated on popular 3D object model databases, the view based approaches \cite{johns2016pairwise, su2015multi} outperform the volumetric ones \cite{maturana2015voxnet, wu20153d}, as reported in the extensive comparison in \cite{qi2016volumetric}. 

\begin{figure*}[t]
  \centerline{
  \includegraphics[scale=0.17]{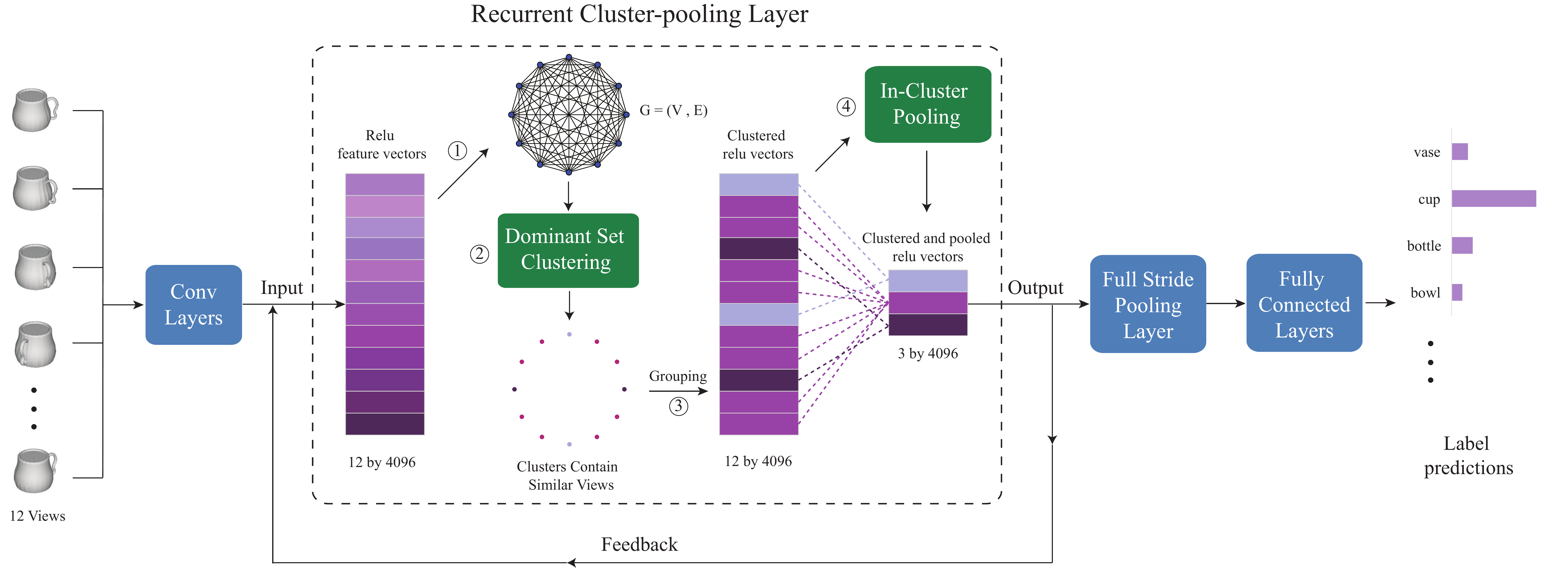}
  }
  \vspace{0.4cm}
  \caption{Our recurrent clustering and pooling CNN. We insert our recurrent cluster-pooling layer (dashed box) after the relu layers (relu6 and relu7 are typical choices) of an Imagenet pretrained VGG-m network. A set of multi-view 2D feature maps (12 views in this example), obtained by projection of the 3D mesh model, is used as input to our system. The proposed layer takes relu vectors of the previous layer as its input and performs
dominant set clustering on a view similarity graph followed by within cluster pooling in a recurrent manner. Once the clusters are stable the outputs serve as inputs to a full stride pooling layer, following which the multi-view relu vectors (12 by 4096) become one unified vector (1 by 4096), which is processed by fully connected layers to produce predicted object category labels.
}
  \label{fig:SystemArchit}
\end{figure*}

The state-of-the-art view based methods differ in the manner in which multi-view information is fused. In the MVCNN approach \cite{su2015multi}, a view pooling layer is inserted in a VGG-m style network to perform multi-view feature fusion of CNN relu vectors. This view pooling layer performs a full stride channel-wise max pooling to acquire a unified feature vector, following which fully connected layers are used to predict category labels. In the pairwise decomposition method in \cite{johns2016pairwise}, two CNNs are used, one for view pair selection and one for pairwise label prediction. The two CNNs each use a VGG-m structure \cite{chatfield2014return} but they have to be trained separately, which is costly. At the expense of increased training cost, the pairwise formulation out performs the MVCNN approach.

In the present article we propose a revised architecture which aims to overcome potential limitations of the above two strategies, namely: 1) the winner-take-all pooling strategy in \cite{su2015multi}, which could discard information from possibly informative views, and 2) the pairwise formulation of \cite{johns2016pairwise}. The key contribution is the introduction of a recurrent clustering and pooling module based on the concept of dominant sets, as illustrated in Figure (\ref{fig:SystemArchit}). The 2D views of an object are abstracted into relu vectors, which serve as inputs to this new layer. Within this layer we construct a view similarity graph, whose nodes are feature vectors corresponding to views and whose edge weights are pairwise view similarities. We then find dominant sets within this graph, which exhibit high within cluster similarity and between cluster dissimilarity. Finally, we carry out channel wise pooling but only from {\em within} each dominant set. If the dominant sets have changed from the previous iteration, they are fed back as new feature vectors to the same layer, and the clustering and pooling process is repeated. But if not, they are fed forward to the next full stride pooling layer.
In contrast to the MVCNN approach of \cite{su2015multi} we only pool information across similar views. The recurrent nature allows for the feature vectors themselves to be iteratively refined. Our recurrent cluster-pooling layer, followed by a full stride view pooling layer, can be inserted after a pretrained network's relu layers to yield a unified multi-view network which can be trained in an end-to-end manner.\footnote{Note that the within-cluster pooling operation is fixed: it is either max or average pooling throughout the recurrence.}
Our experimental results in Section (\ref{sec:EXP}) show that when inserted before the view pooling layer in the MVCNN architecture of \cite{su2015multi}, our recurrent clustering and pooling unit greatly boosts performance, achieving new state-of-the-art results in multi-view 3D object recognition on the ModelNet40 dataset.

\section{Recurrent Clustering and Pooling Layer}\label{sec:RClusterPool_layer}

The recurrent clustering and pooling layer requires the construction of a view similarity graph, the clustering of nodes (views) within this graph based on dominant sets, and the pooling of information from within each cluster. We now discuss these steps in greater depth and provide implementation details.

\subsection{A View Similarity Graph}\label{sec:similarity}
First, a pairwise view similarity measure is defined for any two views in the set of rendered views of a 3D object. We then construct a view similarity graph $G = (V , E , w)$ where views $i, j \in V$ are distinct nodes and each edge $E(i,j)$ has a weight $w(i,j)$ corresponding to the similarity between the views $i$ and $j$. This results in a complete weighted undirected graph $G = K_n$, where $n$ is the number of views.

Different rendered views have different appearances, which are in turn captured with low dimensional signatures by the relu features from a CNN (we typically apply relu6 or relu7 vectors). A very convenient notion of pairwise view \textit{similarity} between the appearance images of views $i$ and $j$ is therefore given by the inner product of the corresponding CNN relu feature vectors $r_i$ and $r_j$:  
\begin{equation}
w(i,j) = r_i \cdot r_j.
\end{equation}
We exploit the property that the components of relu feature vectors from a CNN, where relu stands for ``rectified linear units'', are non-negative and have finite values that tend to lie within a certain range. The larger $w(i,j)$ is, the more similar the two views $i$ and $j$ are.

\subsection{Dominant Set Clustering}\label{sec:DS}
We now cluster views within the view similarity graph, based on the concept of dominant sets \cite{PavPelCVPR2003,PavPel07}. The views to be clustered are represented as an undirected edge-weighted graph with no self-loops $G = (V, E,w)$, where $V = \{1, . . . , n\}$ is the vertex set, $E \subseteq V \times V$ is the edge set, and $w : E \rightarrow \mathbb{R}_+^*$ is the (positive) weight function. As explained in Section (\ref{sec:similarity}), the vertices in $G$ correspond to relu vectors abstracted from different rendered views of a given object, the edges represent view relationships between all possible view pairs, and the edge-weights encode similarity between pairs of views. We compute the affinity matrix of $G$, which is the $n \times n$ nonnegative, symmetric matrix $A = (a_{ij})$ with entries $a_{ij} = w(i, j)$. Since in $G$ there are no self-loops, all entries on the main diagonal of $A$ are zero. 

For a non-empty subset $S \subseteq V$, $i \in S$, and $j \notin S$, let us define
\begin{equation}
\label{eq1}
\phi_S(i,j)=a_{ij}-\frac{1}{|S|} \sum_{k \in S} a_{ik}
\end{equation}
which measures the (relative) similarity between vertices $j$ and $i$, with respect to the average similarity between $i$ and its neighbors in $S$. Next, to each vertex $i \in S$ we assign a weight defined (recursively) as follows:
\begin{equation}
w_S(i)=
\begin{cases}
1,&\text{if\quad $|S|=1$},\\
\sum_{j \in S \setminus \{i\}} \phi_{S \setminus \{i\}}(j,i)w_{S \setminus \{i\}}(j),&\text{otherwise}.
\end{cases}
\end{equation}
As explained in \cite{PavPelCVPR2003,PavPel07}, a positive $w_S(i)$ indicates that adding $i$ to the elements in $S$ will increase its internal coherence, whereas a negative weight indicates that adding $i$ will cause the overall coherence of $S$ to be decreased. Finally, we define the total weight of $S$
\begin{equation}
W(S)=\sum_{i \in S}w_S(i)~.
\end{equation}
\begin{definition}
A non-empty subset of vertices $S \subseteq V$ such that $W(T) > 0$ for any non-empty $T \subseteq S$, is said to be a {\em dominant set} if:
\begin{enumerate}

\item  $w_S(i)>0$, for all $i \in S$,
\item  $w_{S \cup \{i\}}(i)<0$, for all $i \notin S$.
\end{enumerate}
\end{definition}
It is evident from its definition that a dominant set satisfies the two properties of a cluster that we desire, namely, it yields a partition of $G$ with a high degree of intra-cluster similarity and inter-cluster dissimilarity. Condition 1 indicates that a dominant set is internally coherent, while condition 2 implies that this coherence will be decreased by the addition of any other vertex. 
A simple and effective optimization algorithm to extract a dominant set from a graph based on the use of {\em replicator dynamics} can be found in \cite{PavPelCVPR2003,PavPel07,bulo2017dominant}, with a run time complexity of $\mathcal{O}(V^2)$, where $V$ is the number of vertices in the graph. We adopt this algorithm in our implementation.

\begin{figure*}[t]
  \centerline{
  \includegraphics[scale=0.17]{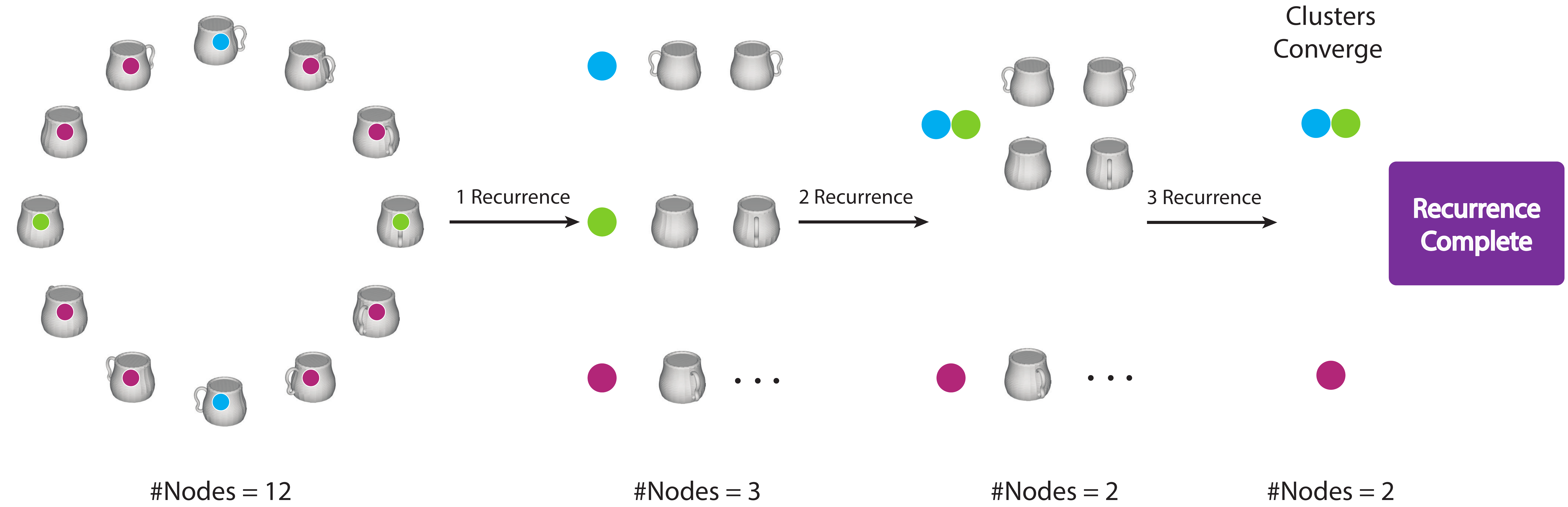}
  }
  \vspace{0.4cm}
  \caption{We provide an illustration of the clustering and pooling recurrences that occur for the case of a mug with 12 initial views. See text for a discussion.}
  \label{fig:ClusterHierarchy}
\end{figure*}

\subsection{Clustering, Pooling and Recurrence}

After obtaining a partition into dominant sets, we propose to perform pooling operations only {\em within} each cluster. This allows the network to take advantage of informative features from possibly disparate views. The resultant relu vectors are then fed back to the beginning of the clustering and pooling layer, to serve as inputs for the next recurrence. This process is repeated, in a cyclical fashion, until the clusters (dominant sets) do not change. During the recurrences, we alternate max and average pooling, which allows for further abstraction of informative features. A full stride max pooling is applied to the relu vectors corresponding to the final, stable, dominant sets. We carry out experiments to demonstrate the effect of different recurrent structures and pooling combinations in Section (\ref{sec:ResClusterPool}).

Figure (\ref{fig:ClusterHierarchy}) depicts the cluster-pooling recurrences for a mug, with 12 initial input views. After the first stage there are 3 clusters, one corresponding to side views (blue), a second corresponding to front or rear views (green) and a third corresponding to oblique views (maroon). At this stage the three relu feature vectors represent pooled information from within these distinct view types. After the second recurrence, information from the 
green and blue views is combined, following which no additional clusters are formed.

The proposed recurrent clustering and pooling strategy aims to improve discrimination between different input objects. When input relu features are changed, our method increases the likelihood of variations in the aggregated multi-view relu output. In contrast, a single max pooling operation, as in the MVCNN approach of \cite{su2015multi}, will result in the same output, unless the change in the inputs cause the present max value to be surpassed.
The ability to capture subtle changes in relu space helps in discrimination between similar object categories. Dominant set clustering seeks to prevent outliers in a given cluster, and the within cluster pooling decreases the likelihood that the aggregated result is determined by the single relu vector that gives the maximum response.

\subsection{Back propagation}
\begin{table}[h]
\centering
    \begin{minipage}{.4\linewidth}
	\centering
	     \includegraphics[scale=0.8]{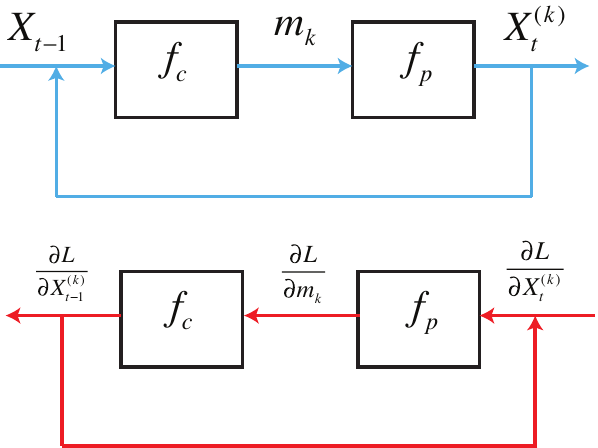}
    \end{minipage}%
    \hfill
	\begin{minipage}{.55\linewidth}
	\centering
	     \includegraphics[scale=0.75]{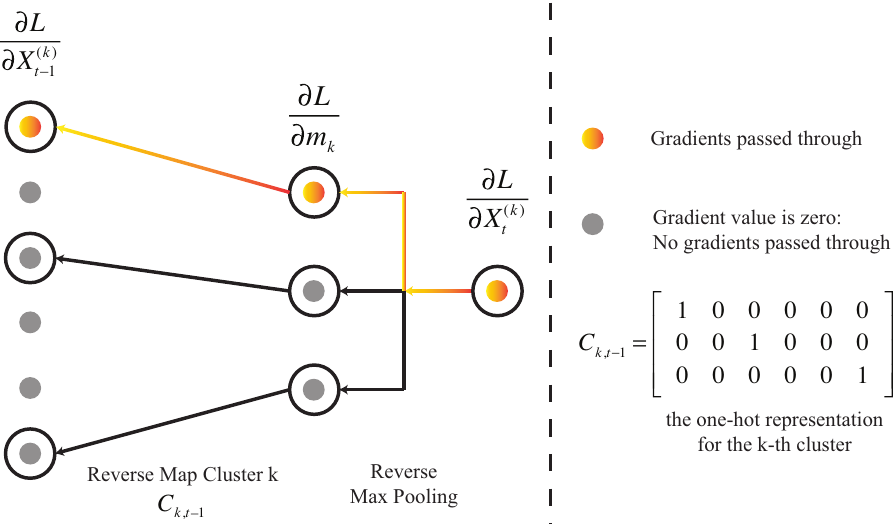}
	\end{minipage} 
	\begin{minipage}[t]{.4\linewidth}

	\captionof{figure}{The recurrent clustering and pooling layer's forward and backward pass with respect to the k-th cluster. Recurrence: (t-1).}
	\label{fig:fp_bp}
	\end{minipage}
	\hfill
	\begin{minipage}[t]{.55\linewidth}
	\captionof{figure}{A toy example with 6 input views. The back propagation for the k-th cluster output with a max within-cluster pooling is illustrated in this example.}
	\label{fig:toyexample}
	\end{minipage}
	
\end{table}

There are no parameters to be learned in our recurrent clustering and pooling layer, therefore we aim to derive the gradients w.r.t this layers' input given gradients w.r.t output to facilitate learning at preceding layers. Details of the forward pass and back propagation steps are illustrated in Figure (\ref{fig:fp_bp}). Here $f_c$ represents the dominant set clustering unit which takes as input the (t-1)-th recurrence $X_{t-1} \in \mathbb{R}^{n_{t-1} \times d}$ and outputs cluster assignments, where $n_{t-1}$ is the number of input nodes and $d$ is relu vector's dimension. $f_p$ stands for the within-cluster pooling unit, which takes as input relu vectors belonging to the k-th resultant cluster $m_k \in \mathbb{R}^{c_k \times d}$ and outputs a pooled relu vector, where $c_k$ is the number of nodes in k-th cluster. During the \textit{forward} pass, we first acquire cluster assignments for the (t-1)-th recurrence using the dominant set algorithm 
$C_{k,t-1} = f_c \left( A_{t-1} \right)$, 
where $k$ stands for an arbitrary resultant cluster and $A_{t-1}$ stands for the affinity matrix of the constructed similarity graph. A one hot cluster representation matrix $C_{k,t-1} \in \mathbb{R}^{c_k \times n_{t-1}}$ is constructed in the following manner. For an arbitrary cluster $k$ containing input nodes $\{k_1,k_2,k_3,..., k_{c_k}\}$, the i-th row in its cluster representation matrix is a one-hot vector encoding value $k_i$. Now, inputs belonging to the k-th cluster can be represented as 
\begin{equation}\label{eqn:m_k}
m_k = C_{\left(k,t-1\right)} X_{t-1}.
\end{equation}
The within-cluster pooling unit will then give
$X_t^{\left(k\right)} = f_p \left( m_k \right)$, 
where $X_t^{\left(k\right)} \in \mathbb{R}^{1 \times d}$ is the pooled relu vector of cluster k. The above process applies to all resultant clusters.

To establish the formulas for \textit{back propagation} of the recurrent clustering and pooling layer, we define the cross entropy loss functions as $L$. We note that the backward pass requires the same amount of recurrence as the forward pass, but the direction of data flow is opposite as shown in Figure (\ref{fig:fp_bp}). During the backward pass of (t-1)-th recurrence, we iteratively loop through clusters to accumulate the gradients of the loss function w.r.t the (t-1)-th recurrence's input defined as $\frac{\partial L}{\partial X_{t-1}}$. For a given resultant cluster $k$, gradients will be mapped back only to those input nodes that belong to this very cluster. If we define the gradients w.r.t the recurrent layer's inputs of cluster k as $\frac{\partial L}{\partial X_{t-1}^{\left(k\right)}}$, gradients w.r.t output as $\frac{\partial L}{\partial X_{t}^{\left(k\right)}}$, and w.r.t pre-pooling as $\frac{\partial L}{\partial m_k}$, we have the following equations:

\begin{subequations}
\begin{minipage}{.45\linewidth}
\begin{equation}
\frac{\partial L}{\partial m_k} = f_p^{-1} \left( \frac{\partial L}{\partial X_{t}^{\left(k\right)}} \right)
\end{equation}
\end{minipage}
\hfill
\begin{minipage}{.45\linewidth}
\begin{equation}
\frac{\partial L}{\partial X_{t-1}^{\left(k\right)}}  = C_{k,t-1}^{T} \frac{\partial L}{\partial m_k}
\end{equation}
\end{minipage}
\end{subequations}
which are derived by reversing the operations of pooling and then clustering. Gradients with respect to the recurrent layer's input are then given by
\noindent
\begin{equation}\label{eqn:gradientFinal}
\frac{\partial L}{\partial X_{t-1}} =  \sum_{k} \frac{\partial L}{\partial X_{t-1}^{\left(k\right)}} = \sum_{k} C_{k,t-1}^{T} f_p^{-1} \left( \frac{\partial L}{\partial X_{t}^{\left(k\right)}} \right).
\end{equation}

The toy example in Figure (\ref{fig:toyexample}) illustrates a scenario with 6 input views using within-cluster max pooling. When back propagating the gradients, the orange cells represent the relu positions where gradients have passed through. We note that the grey cells' gradients will always remain zero and gradients w.r.t input $X_{t-1}$ will only have non-zero values at nodes belongs to the k-th cluster.

\newcommand{\featrow}[6]{#1 & #2 & #3 & #4  & #5  & #6 \\}
\newcommand\featFigScale{0.25}
\begin{table*}[t]
\scalebox{0.9}{
\begin{tabular}{cccccc}
\featrow{cup}{chair}{guitar}{lamp}{plant}{toilet}
\featrow{\includegraphics[scale=\featFigScale]{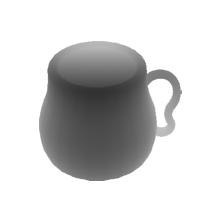}}{\includegraphics[scale=\featFigScale]{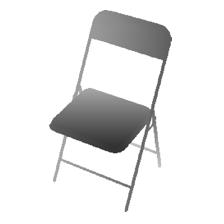}}{\includegraphics[scale=\featFigScale]{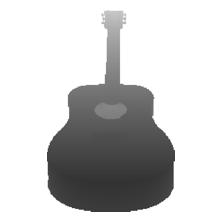}}{\includegraphics[scale=\featFigScale]{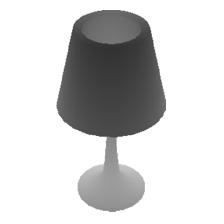}}{\includegraphics[scale=\featFigScale]{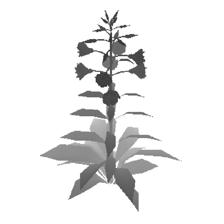}}{\includegraphics[scale=\featFigScale]{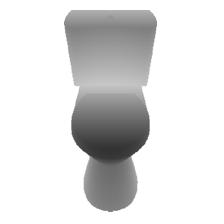}}
\featrow{\includegraphics[scale=\featFigScale]{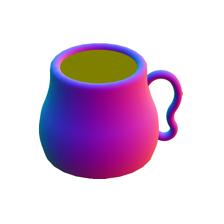}}{\includegraphics[scale=\featFigScale]{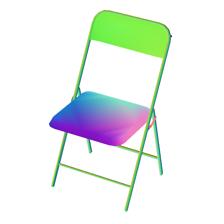}}{\includegraphics[scale=\featFigScale]{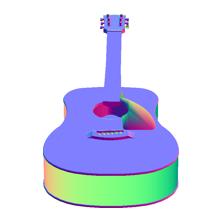}}{\includegraphics[scale=\featFigScale]{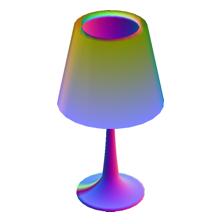}}{\includegraphics[scale=\featFigScale]{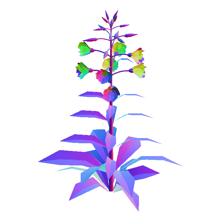}}{\includegraphics[scale=\featFigScale]{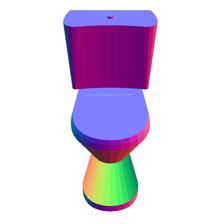}}
\end{tabular}
}
\vspace{0.2cm}
\captionof{figure}{Rendered views of ModelNet40 objects using additional feature modalities. Top row: depth map. Bottom row: surface normals rendered with an RGB colormap, with green pointing towards the viewer, blue pointing upwards and red pointing right.}
\label{fig:featTypes}
\end{table*}

\section{Experiments}\label{sec:EXP}
\textbf{Network Setup.} We use the same baseline CNN structure as in \cite{johns2016pairwise,su2015multi}, which is a VGG-M network \cite{chatfield2014return} containing five convolutional layers followed by three fully connected layers. We follow the same process of network pretraining and task specific network fine tuning as in \cite{johns2016pairwise,qi2016volumetric,su2015multi}. Specifically, we use the Imagenet 1k pretrained VGG-m network provided in \cite{chatfield2014return} and fine tune it on the ModelNet 40 training set after our recurrent clustering and pooling layer is inserted. In our experiments, we insert customized layers after the relu6 layer. 

\medskip
\noindent
\textbf{Dataset.} We evaluate our method on the Princeton ModelNet40 dataset \cite{modelnet} which contains 12, 311 3D CAD models from 40 categories. This dataset is well annotated and many state-of-the-art approaches have reported their results on it \cite{johns2016pairwise, qi2016volumetric, su2015multi}. The dataset also provides a training and testing split, in which there are 9,843 training and 2,468 test models \footnote{Qi \etal \cite{qi2016volumetric} used this entire train/test split and reported average class accuracy on the 2,468 test objects. Su \etal \cite{su2015multi} used a subset of train/test split comprising the first 80 objects in each category in the train folder (or all objects if there are fewer than 80) and the first 20 objects of each category in the test folder, respectively.}. We use the entire training and testing set for experiments in Sections (\ref{sec:ResClusterPool}) and we provide results for both the full set and the subset when comparing against related works in Section (\ref{sec:ExpArt}). 

\medskip
\noindent
\textbf{Rendering and additional feature modalities.} We render the 3D mesh models by placing 12 centroid pointing virtual cameras around the mesh every 30 degrees with an elevation of 30 degrees from the ground plane, which is the first camera setup in \cite{su2015multi}. In our experiments we also include additional feature types beyond the grey scale (appearance) images, to encode surface geometry. Surface normals are computed at the vertices of each 3D mesh model and then bilinearly interpolated to infer values on their faces. For depth, we directly apply the normalized depth values. We then render these 3D mesh features using the multi-view representations in \cite{su2015multi} but with these new features. We linearly map the surface normal vector field $(n_x,n_y,n_z)$ where $n_i \in [-1,1]$ and $ i = x,y,z$, to a color coding space $(C_x, C_y, C_z)$ where $C_i \in [0,255]$ for all $i = x,y,z$, to ensure that these features are similar in magnitude to the intensity values in the grey scale appearance images. Examples of the computed rendered feature types are shown in Figure (\ref{fig:featTypes}).

\subsection{Training and Testing Procedure}\label{sec:Training}
We explore two training approaches for our system in Figure (\ref{fig:SystemArchit}). In {\em Fast training}
we use the Imagenet pretrained VGG-m network in \cite{chatfield2014return} to compute the relu7 feature vectors of each view of each training 3D mesh, by forward passing the 2D views into it. We then forward pass the relu7 vectors to our recurrent clustering and pooling layer. 
In our experiments we use a universal clustering scheme at each recurrence for all training and testing objects, by averaging over the affinity matrices of all training objects. Ideally category-specific clustering is preferred for better category level recognition accuracy, but we do not have access to labels at test time. The universal clustering scheme is computationally efficient and the consistency it grants helps improve recognition accuracy. We record each recurrence's universal clustering scheme w.r.t training objects to formulate a recurrent clustering hierarchy for end-to-end training.
After the full stride pooling layer, we have a single fused relu7 feature vector on which an SVM classifier is trained. At test time, we follow the same routine for a test object and apply the clustering scheme we used during training. The trained SVM classifier is then applied to predict its object category label. We note that there is no CNN training at all since we applied an Imagenet pretrained VGG-m network with no fine-tuning. In {\em End-to-end training} we directly feed in rendered 2D maps of training objects to the unified network in Figure (\ref{fig:SystemArchit}) to perform the forward and backward passes in an end-to-end manner, using the recorded recurrent clustering hierarchy. The weights for the VGG-m network's layers before and after the recurrent clustering and pooling layer are jointly learned during training time. At test time, we send a test object's rendered 2D maps to the network to acquire its predicted object category label.

\begin{table}[t]
\renewcommand{\arraystretch}{1.25}
    \begin{minipage}{.5\linewidth}
      \centering
	\scalebox{0.7}{
	\begin{tabular}{C{4cm}|C{1.6cm}C{1.7cm}C{1.5cm}}
	\hline
	Cluster Pooling Structure            & Fast & End-to-End \\
	\hline
	mvcnn (f-max)                             & 89.8                  & 91.5                    \\
    DS-avg-f-max                        & 90.4                & 91.1                 \\
	DS-max-f-avg                        & 91.2                  &  91.3                 \\
	\rowcolor{pink}(DS-alt)-f-max            & 91.9                 & 92.2                    \\

	\hline 
	\end{tabular}
	
	}
    \end{minipage}%
    \begin{minipage}{.5\linewidth}
      \centering
	\scalebox{0.7}{
	\begin{tabular}{C{1.2cm}C{1.2cm}C{1.2cm}C{1.5cm}}
	\hline
	RGB & Depth & Surf  & Accuracy \\
	\hline 
	\cmark       & \xmark    & \xmark                     & 91.9     \\
	\cmark       & \cmark   & \xmark                      & 92.9     \\
	\cmark       & \xmark   & \cmark                      & 92.1     \\
	\xmark       & \cmark   & \cmark                      & 92.9     \\
	\cmark       & \cmark   & \cmark                     &  \fbox{93.3}   \\
	\hline
	\end{tabular}
	
	}
    \end{minipage} 
    \begin{minipage}[b]{.45\linewidth}
	\caption{Comparison of different pooling structures and training types.  }
	\label{tab:ResClusterPooling}
	\end{minipage}
	\hfill
	\begin{minipage}[b]{.45\linewidth}
	\caption{The benefits of additional feature modalities (see text).}
	\label{tab:ResFeatures}
	\end{minipage}
\end{table}

\subsection{Model Ablation Study}
\paragraph{Recurrent Clustering and Pooling Structure} \label{sec:ResClusterPool}
In our evaluation ``f-max'' stands for a full stride channel-wise max pooling, ``ds-avg'' stands for one recurrence of the clustering and pooling layer with within-cluster average pooling and ``(ds-x)'' stands for recurrent clustering and pooling until the clusters are stable. We examine the benefits of both recurrence and pooling with 3 variations: 1) ds-avg-f-max, 2) ds-max-f-avg, which uses only one phase of clustering and pooling but with different pooling operations and 3) (ds-alt)-f-max which uses recurrent clustering while alternating max and average pooling. 
Table (\ref{tab:ResClusterPooling}) shows the results of these variations, together with the baseline ``f-max'' used by the MVCNN method \cite{su2015multi}. Recurrent clustering and pooling is indeed better than a non-recurrent version and alternating max-avg within-cluster pooling followed with a full stride max pooling performs better than the other variants. We further note that end-to-end training performs better than non-end-to-end fast training.

\vspace{-0.2cm}
\paragraph{Additional Feature Types}\label{sec:ExpFeat}
We now explore the benefit of additional feature modalities using ``(ds-alt)-f-max'' clustering and pooling. We run experiments using the fast training scheme introduced in section (\ref{sec:Training}). The results in Table (\ref{tab:ResFeatures}) show that even without fine tuning, the network pretrained on pure appearance images can be generalized to handle different feature modalities. The additional feature types
significantly boost the recognition accuracy of our recurrent cluster and pooling structure with a combination of appearance, depth and surface normals giving the best test set recognition accuracy of 93.3\% with no CNN training.

\paragraph{Effects from number of Views}\label{sec:ExpNViews}
\begin{table}[t]
\centering
    \begin{minipage}{.5\linewidth}
    \scalebox{0.7}{
    \renewcommand{\arraystretch}{1.3}
		\begin{tabular}{C{2cm}|C{.8cm}C{.8cm}C{.8cm}C{.8cm}C{.8cm}}
		\hline
		Cluster Pooling Structure             & End-to-End? & 3 Views & 6 Views & 12 Views & 24 Views \\
		\hline
		mvcnn                          &  \xmark        &    90.76         &    91.16          &    89.8            &       91.93              \\
	    \rowcolor{cyan}mvcnn                           & \cmark         &      91.33       &     92.01         &        91.49        &        92.42             \\
		Ours                & \xmark        &     91.45         &    91.61          &    91.89            &      92.70  \\
		\rowcolor{orange}Ours                 & \cmark        &      92.1        &      92.22        &        92.18        &         93.0            \\
	
		\hline 
		\end{tabular}
	}
        
    \end{minipage}%
    \hfill
	\begin{minipage}{.5\linewidth}
	\centering
	     \includegraphics[scale=0.25]{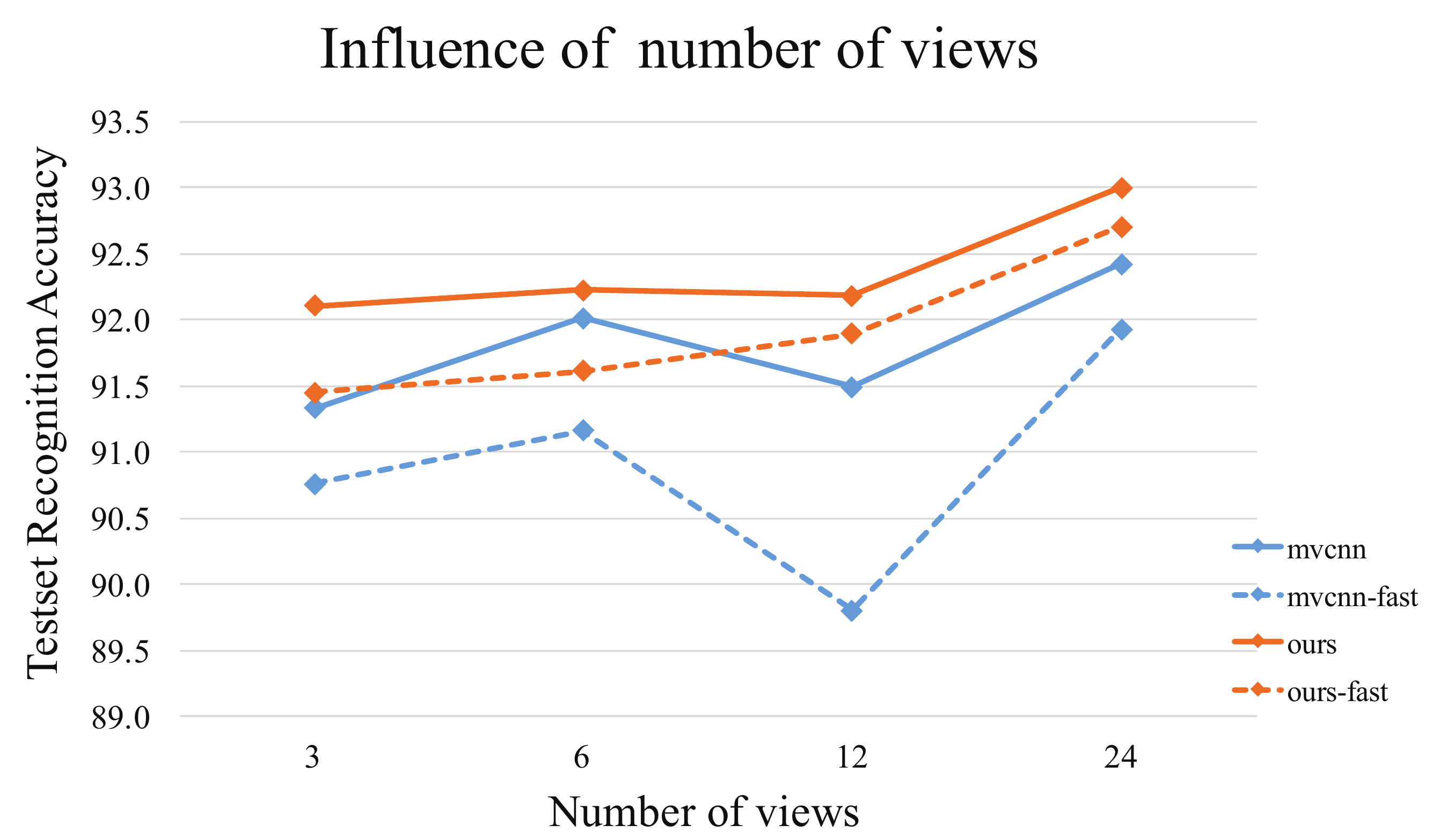}
	\end{minipage} 
	\begin{minipage}[b]{.45\linewidth}
	\caption{Effect of the number of views.}
	\label{tab:nViewsExp}
	\end{minipage}
	\hfill
	\begin{minipage}[b]{.45\linewidth}
	\captionof{figure}{Effect of the number of views.}
	\label{fig:nViewsExp}
	\end{minipage}
	
\end{table}

We evaluate the effect of the number of views on our recurrent clustering and pooling CNN and on MVCNN \cite{su2015multi} in Table (\ref{tab:nViewsExp}) and Figure (\ref{fig:nViewsExp}). Our method is consistently better than the MVCNN approach, with a steady improvement of recognition accuracy as the number of views increases. We further note that for MVCNN there is an evident drop in performance when increasing from 6 views to 12 views, illustrating a potential drawback of its single full stride max pooling strategy.

\subsection{Comparison with the present State-of-the-art}\label{sec:ExpArt}

We now compare our method against the state-of-the-art view based 3D object recognition approaches \cite{johns2016pairwise, qi2016volumetric, su2015multi} in Table (\ref{tab:CompStateArt}). Two accuracies groups are reported: one for the subset used in MVCNN \cite{su2015multi} and one for the full set used in \cite{qi2016volumetric,johns2016pairwise}.\footnote{Judging by the description in \cite{johns2016pairwise} ``ModelNet10, containing 10 object categories with 4,905 unique objects, and ModelNet40, containing 40 object categories and 12,311 unique objects, both with a testing-training split.'' we assume they used the entire train/test split. Note that 9, 843 training and 2, 468 test models result in a total of 12,311 objects in the dataset.} For Johns \etal \cite{johns2016pairwise} and Qi \etal \cite{qi2016volumetric}, we quote the recognition accuracies reported in their papers. For MVCNN \cite{su2015multi}, we quote their reported subset results but reproduce experimental results on the full set.  In \cite{johns2016pairwise, qi2016volumetric, su2015multi}, two types of view sampling/selection strategies are applied, as mentioned in Table (\ref{tab:CompStateArt}).\footnote{
The term ``$30^{\circ}$ elevation'' means only views at an elevation of $30^{\circ}$ above the ground plane, constrained to rotations about a gravity vector, are selected. The term ``Uniform" means all uniformly sampled view locations on the view sphere. More specifically in Johns \etal \cite{johns2016pairwise} where a CNN based next-best-view prediction is applied for view point selection, the authors select the best 12 over 144 views to perform a combined-voting style classification. Therefore their CNNs are trained on a view base of 144 rendered views per object.} 

In addition to recognition accuracy, we also consider training time consumption. We measure the training time cost per epoch by 
$\mathcal{C} = \Phi \mathcal{N}_v \mathcal{N}_c$
where $\mathcal{N}_v$ denotes the number of rendered views, $\mathcal{N}_c$ denotes the number of VGG-m like CNNs trained and $\Phi$ stands for, for a single CNN, the \textit{unit training cost} in terms of computational complexity for one epoch over a given 3D dataset when only one view is rendered per object. We denote the computational complexity for our fast training scheme, which consists of 1 epoch of forward passing and SVM training, as $\epsilon$, since this is less costly than $\Phi$. We estimate that $\epsilon < 0.5\Phi$. 

The results in Table (\ref{tab:CompStateArt}) show that our recurrent clustering and pooling CNN out performs the state-of-the-art methods in terms of recognition accuracy, achieving a 93.8 \% full test set accuracy on ModelNet40 and 92.8\% on the subset. When fast (non end-to-end) training is applied, the method still achieves state-of-the-art recognition accuracies of 93.3\% and 92.1\% with a greatly reduced training cost. Our results presently rank second on the ModelNet40 Benchmark \cite{modelnet}.\footnote{As of July 17th, 2017.} The top performer is the method of Brock \etal \cite{brock2016generative}, which is a voxel-based approach with 3D representation training, focused on designing network models. In contrast, our method is a 2D view-based approach which exploits strategies for view feature aggregation. As rightly observed by a reviewer of this article, the 95.54\% accuracy on ModelNet40 achieved in \cite{brock2016generative} is accomplished with an ensemble of 5 Voxception-ResNet \cite{he2016deep} (VRN) architecture (45 layers each) and 1 Inception \cite{szegedy2017inception} like architecture. Training each model from the ensemble takes 6 days on a Titan X. In our approach, fine-tuning a pretrained VGG-M model after inserting our recurrent layer takes 20 hours on a Tesla K40 (rendering 12 views for ModelNet40). When only one VRN is used instead of the ensemble, Brock \etal achieve 91.33\% on ModelNet40, while we achieve 92.2\% using only RGB features.

\newcommand{\resultrow}[7]{#1 & #2 & #3 & #4  & #5  & #6 & #7\\}
\begin{table*}[t]
\centering
\renewcommand{\arraystretch}{1.35}
\scalebox{0.65}{
\begin{tabular}{L{2.8cm}|C{1.4cm}C{1.6cm}C{2.3cm}C{1.6cm}C{2.0cm}C{1.5cm}C{1.5cm}}
\hline
Method          & View Selection & \# Views & Feature Types                                                         & base CNNs  & Training Cost per Epoch & Subset Accuracy  & Fullset Accuracy\\
\hline
\multirow{2}{*}{Pairwise \cite{johns2016pairwise}  }    
& \resultrow{$30^{\circ}$}{best 12 of 144 }{RGB}{2x vgg-m}{\cellcolor{blue!25} $288 \Phi$ }{n/a}{90.7}
& \resultrow{uniform}{best 12 of 144 }{RGB + Depth}{4x vgg-m}{\cellcolor{blue!25}$576 \Phi$}{n/a}{ 92.0}
\hline
Qi-MVCNN \cite{qi2016volumetric} 
& \resultrow{uniform}{20 }{RGB + Sph-30 + Sph-60 }{3x alexnet}{$60  \phi = 5 \Phi $}{n/a}{91.4}
\Xhline{0.01cm}
{Su-MVCNN \cite{su2015multi} }          
& \resultrow{$30^{\circ}$}{12 }{RGB}{1x vgg-m}{$12 \Phi$ }{89.9}{91.5}
\Xhline{0.03cm}
\multirow{2}{*}{Ours-Fast}
& \resultrow{$30^{\circ}$ }{12 }{RGB }{1x vgg-m}{$ \epsilon < 0.5\Phi$}{90.4}{91.9}
& \resultrow{$30^{\circ}$ }{12 }{RGB + Depth + Surf  }{1x vgg-m}{$3 \epsilon < 1.5\Phi$}{\cellcolor{red!25}92.1}{\cellcolor{red!25}93.3}
\hline
\multirow{2}{*}{Ours-End-To-End}  
& \resultrow{$30^{\circ}$ }{12 }{RGB }{1x vgg-m}{$12 \Phi$}{91.5}{92.2}
& \resultrow{$30^{\circ}$ }{12 }{RGB + Depth + Surf }{3x vgg-m}{$36 \Phi$}{\cellcolor{red!25}92.8}{\cellcolor{red!25}93.8}
\Xhline{0.03cm}
\end{tabular}
}
\vspace{0.3cm}
\caption{A comparison against state-of-the-art view based 3D object recognition methods. For additional feature types, we used the same acronyms as is in Section (\ref{sec:ExpFeat}). For Qi \etal \cite{qi2016volumetric}, Sph-30 and Sph-60 stand for 30 and 60 uniform samples on a sphere. The training cost per epoch provides an estimate of the CNN training time consumption for each method. In \cite{johns2016pairwise}, 2 CNNs need to be trained per feature modality, making it more costly. $\phi$ denotes the unit cost for alexnet where $\Phi \approx 12\phi$, as observed in our experiments.}
\label{tab:CompStateArt}
\end{table*}


\vspace{-0.2cm}
\section{Conclusion}\label{sec:CON}

The recurrent clustering and pooling layer introduced in this paper aims to aggregate multi-view features in a way that provides more discriminative power for 3D object recognition. Experiments on the ModelNet40 benchmark demonstrate that the use of this layer in a standard pretrained network achieves state of the art object category level recognition results. Further, at the cost of sacrificing end-to-end training, it is possible to greatly speed up computation with a negligible loss in multi-view recognition accuracy. We therefore anticipate that the application of a recurrent clustering and pooling layer will find value in 3D computer vision systems in real world environments, where both performance and computational cost have to be considered.

\paragraph{Acknowledgments} We are grateful to the Natural Sciences and Engineering Research Council of Canada for funding, to Stavros Tsogkas and Sven Dickinson for their helpful comments, and to the reviewers whose constructive feedback greatly improved this article.

\bibliography{egbib}
\end{document}